\documentclass[10pt, journal]{IEEEtran}
\usepackage{amsmath,amsfonts}
\usepackage{algorithmic}
\usepackage{algorithm}
\usepackage{array}
\usepackage[caption=false,font=footnotesize,labelfont=rm,textfont=rm]{subfig}
\usepackage{caption}
\usepackage{textcomp}
\usepackage{stfloats}
\usepackage{url}
\usepackage{verbatim}
\usepackage{indentfirst,bm}
\usepackage{graphicx}
\usepackage{pifont}
\usepackage{multirow}
\usepackage{epstopdf}
\usepackage{amsthm}
\usepackage{color}
\usepackage{lineno,hyperref}
\usepackage{booktabs}

\usepackage{cite}

\def\BibTeX{{\rm B\kern-.05em{\sc i\kern-.025em b}\kern-.08em
    T\kern-.1667em\lower.7ex\hbox{E}\kern-.125emX}}
\usepackage{balance}
\begin{document}
\title{Unsupervised Abnormal Stop Detection for Long Distance Coaches with Low-Frequency GPS}

\author{Jiaxin Deng, Junbiao Pang, Jiayu Xu, and Haitao Yu
\IEEEcompsocitemizethanks
{
\IEEEcompsocthanksitem J. Deng, J. Pang and J. Xu are with the School of Information Science and Technology, Beijing University of Technology, Beijing 100124, China (e-mail: \mbox{junbiao\_pang@bjut.edu.cn}).

\IEEEcompsocthanksitem  H. Yu is the Beijing Intelligent Transportation Development Center, Beijing 100161, China (email: yuhaitao@jtw.beijing.gov.cn)
 }
}


\maketitle

\begin{abstract}

In our urban life, long distance coaches supply a convenient yet economic approach to the transportation of the public. One notable problem is to discover the abnormal stop of the coaches due to the important reason, \textit{i.e.}, illegal pick up on the way which possibly endangers the safety of passengers. It has become a pressing issue to detect the coach abnormal stop with low-quality GPS. In this paper, we propose an unsupervised method that helps transportation managers to efficiently discover the Abnormal Stop Detection (ASD) for long distance coaches. Concretely, our method converts the ASD problem into an unsupervised clustering framework in which both the normal stop and the abnormal one are decomposed. Firstly, we propose a stop duration model for the low frequency GPS based on the assumption that a coach changes speed approximately in a linear approach. Secondly, we strip the abnormal stops from the normal stop points by the low rank assumption. The proposed method is conceptually simple yet efficient, by leveraging low rank assumption to handle normal stop points, our approach enables domain experts to discover the ASD for coaches, from a case study motivated by traffic managers. Datset and code are publicly available at: \url{https://github.com/pangjunbiao/IPPs}.

\end{abstract}

\begin{IEEEkeywords}

Abnormal Stop Detection, Long range coach, Low-rank Matrix Decomposition

\end{IEEEkeywords}

\IEEEdisplaynotcompsoctitleabstractindextext
\IEEEpeerreviewmaketitle

\section{INTRODUCTION}\label{sec:intro}

\newtheorem{myobr}{Observation}
\newtheorem{mydef}{Definition}
\newtheorem{mythe}{Theorem}
\newtheorem{mypro}{Proposition}

Long-range coaches, as a common approach of public transportation in China, offering convenient, economical and time-efficient long distance services, compared to railway travel~\cite{VANACKER2020759}~\cite{HASIAK20161706}. Yet, managing long-range coach industry presents unique challenges due to their long distances and the long time operation, distinguishing them from the other transportation systems, \textit{e.g.}, buses, subways or railway. A notable issue within the long-range coach industry is Illegal Passenger Pickups (IPPs) in China, where:
\begin{mydef}[Definition of IPP]
Coaches abnormally stop at the non-designated locations to pick up or drop off passengers and their luggage.
\end{mydef}

Local laws stipulate that passenger coaches must not pick up passengers outside the designated stops without legitimate reasons and must not alter the designated route. The IPPs activity in China tend to bring the several dangers as follows:
\begin{itemize}
\item [1.] \textit{Failure to the real-name ticketing process.} In China, passengers and their luggage must be checked for security to protect passengers from the potential dangers. \textit{e.g.}, flammable items, knifes which pose significant safety risks~\cite{BIELER2022IEEE}.
\item [2.] \textit{Overcrowding problem and passenger's rights.} The tickets sold from stations constrain the number of the passengers in a coach to avoid the overcrowding problem~\cite{DONG2024IEEE}~\cite{BasultoElias2023StrategyAS}, that is a strictly illegal behavior in China. Besides, if traffic accidents occur, the rights of the passenger without tickets would be impaired. Because a ticket usually contains the traffic accident insurance for this travel in China.

\item [3.]\textit{Possibly causing traffic accidents.} When the IPPs activity is occurred, passengers usually pick or drop their luggage at the side of a road in a rush~\cite{BasultoElias2023StrategyAS}. Moreover, coaches usually occupy a lane of a road. The above two problem tends to cause the traffic accidents.
\end{itemize}

\begin{figure}[t!]
  \centering
    \label{fig:abnormal behavior1} 
    \subfloat[A motor in a coach]{\includegraphics[width=.22\textwidth]{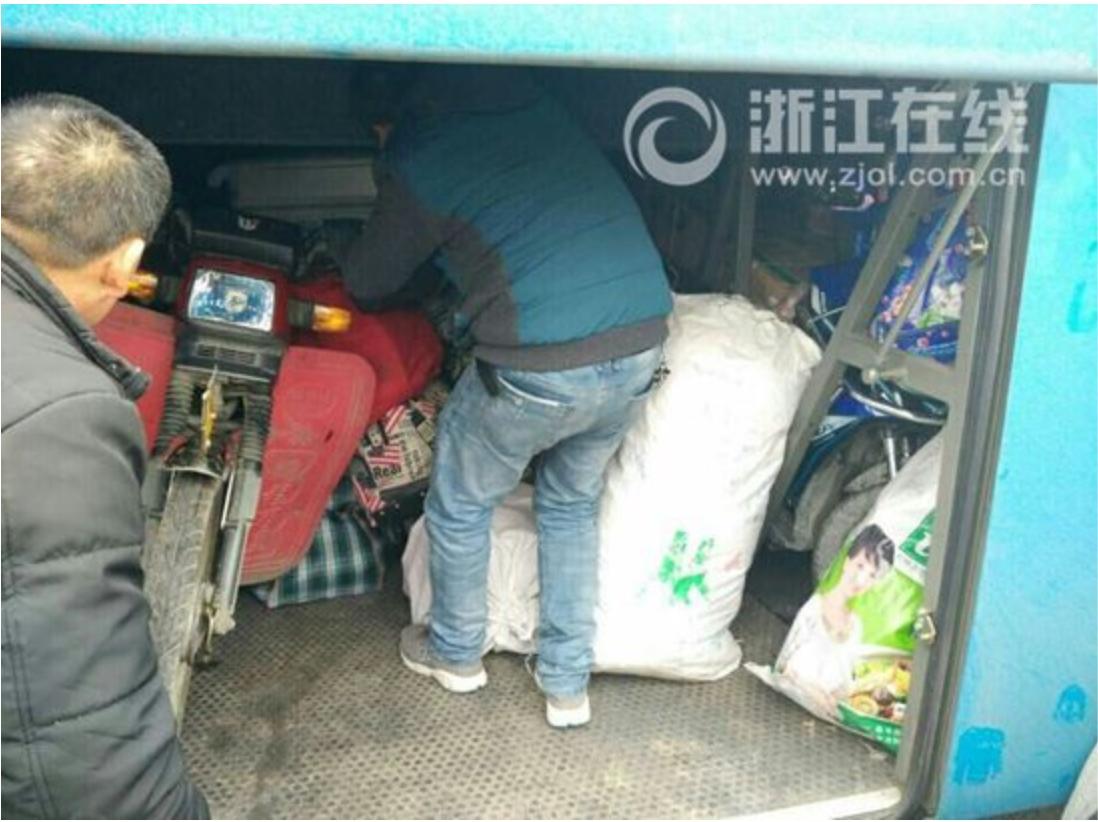}}
    \label{fig:abnormal behavior2} 
    \subfloat[Stop at an intersection]{\includegraphics[width=.25\textwidth]{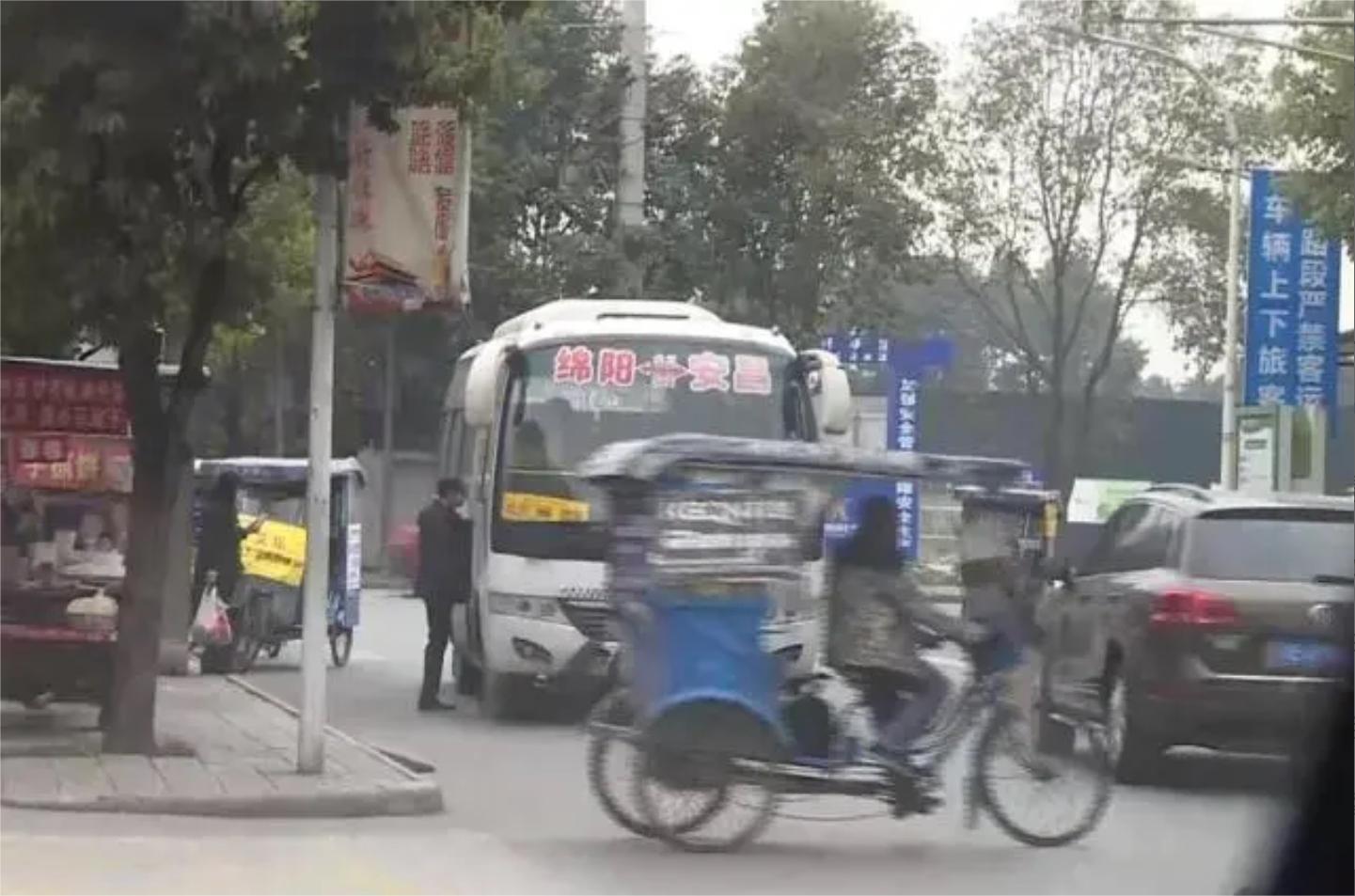}}
  \caption{A case shows that the IPPs activity of long-range coaches tends to carry dangerous luggage (a) and cause traffic accidents (b).\protect\footnotemark}
  \label{fig:The abnormal behavior coaches picking} 
\end{figure}
\footnotetext{\url{https://auto.zjol.com.cn/zjcw/jtjs/201902/t20190220_9494189.shtml}}

The motivation of coach drivers is usually to maximize their income by picking up passengers. In order to instantly prevent IPPs and protect the rights of passengers, transport authority currently checks the correspondence between the number passengers and the ticket records at the fixed inspection stations in China\footnote{\url{https://www.gd.gov.cn/gdywdt/zfjg/content/post_2286796.html}}. This scheme is ineffective since coach drivers could easily bypass the manually check by picking up passengers at a spot along the route with instant messengers, \textit{e.g.}, mobile phones. 
This situation presents a critical issue for the coach industry and the transport authority: \textit{ How can we effectively and precisely identify the spots where IPPs are often occurred in the data-driven way?} Once the spots are identified, the local law enforcers would dynamically check these identified spots to prevent the IPPs activity.

A natural solution is to discover the abnormal stop spots of coaches. Because we have observed that some spots have good traffic conditions for shuttling passengers, \textit{e.g.}, the spot nearby by a bus station. A promising strategy leverages Global Positioning Satellite (GPS) to determine the stop spots (\textit{i.e.}, where the IPPs are occurred) of long-distance coaches. The GPS equipment on each coach in China reveals the locations of stops including the passive stops under the normal scenarios, \textit{e.g.}, congestion, and the red lights at intersections, as well as the active stops \textit{e.g.}, pick up passengers, and avoid an accident. Therefore, in the complex traffic conditions, effectively determining the nature (\textit{e.g.}, passive and active) of the different stop points is a key issue to discover the abnormal stop spots by stripping them from these normal ones. 

\subsection{Research Challenges}

\textit{How to determine the stop spots from the low-frequency GPS:} 
In practice, there exists large amount of low-sampling-rate (\textit{e.g.},
one point every 2 minutes) GPS trajectories for a coach. They are generated in the scenarios where saving of energy cost and communication cost are desired. Low-frequency GPS typically refers to send a GPS coordinate per $30$ seconds in China (see Table~\ref{tab1:Bus GPS data description} for more details). In order to capture the abnormal stop spots, we have to first obtain the possible stop spots during the travel service of a coach. However, a $30$ seconds interval is too long to directly judge whether a coach has stopped or not.

\textit{Discover the abnormal stop spots:}
A coach would have various stop behaviors due to the changes of road conditions and different driving behaviors~\cite{FAN2019607}. 
The abnormal stop spots are the candidate spots where the coach drivers would like to pick up passengers. Therefore, the key challenge of this research is to effectively discover the abnormal stop spots in an unsupervised approach under the complex traffic conditions~\cite{Pangjunbiao2018IEEE}.

\subsection{Our Contribution}

In this paper, we propose a method that aims to resolve the above challenges in an unsupervised approach. Rather than seeking the stop spots by manually screening the physical constraints that conventionally pick up passengers, we instead of directly discover the abnormal stop spots by assuming that the complex traffic system follows the low rank constraints~\cite{vicent24-nature-pyhsica}. That is, we cast our quest for ``abnormal'' stops as the matrix decomposition problem, in which the low rank models the normal stops, while a sparsity residual term represents the abnormal ones. 

This paper makes several significant contributions to the field of the long range coach industry management, as outlined below: 

\textbf{Introduction to the IPPs Problem.} This study is pioneering in exploring the discovery of the spots of the IPPs activity for long range coaches, only relying on the low-frequency GPS. We approach this challenge as an unsupervised problem, achieving the effective solutions for identifying the candidate spots where the IPPs activities tend to occur.

\textbf{Discovery of the possible stop spots from the low-frequency GPS:} 
This paper addresses the problem of the low-frequency GPS in particular. 
We propose a method to identify the possible stop spots by simply assuming that the speed of a coach changes linearly. These candidate stop spots are instrumental in characterizing the stop spots from diverse coach drivers.

\textbf{Determining anomalous stop spots:} 
We assume that the abnormal stop spots are sparse, compared with the normal ones. Besides, it is impossible for long range coaches to illegally stop at some special spots (\textit{e.g.}, main road). Therefore, we introduced the structured sparsity~\cite{Zhang2023UAF} and the low rank constraint~\cite{CAO201710} for detecting abnormal stop spots. Our
findings reveal the proposed method as the superior method for detecting the spots for the IPPs activities.


\section{Related Work }~\label{sec:relatedwork}
To our knowledge, this paper first leverages GPS points to identify the stop spot of the IPPs activity for long range coaches. Therefore, we briefly review the related works about exploiting GPS to discover abnormal spots for the other tasks.

\subsection{Abnormal detection from GPS}

The varying natural environments and traffic conditions~\cite{Yuan2020Electro}~\cite{Pang2019GPS}~\cite{HanShuang2020IET} leverages GPS to address multiple issues related to traffic safety and efficiency. For instance, by monitoring the movement patterns and speed variations of buses, abnormal driving behaviors can be detected, \textit{e.g.}, abnormal driving behaviors~\cite{Boateng2023osti}~\cite{Yu2022ADE}, over speeding~\cite{Zhang2023DatadrivenSM}, the deviation from the predetermined route~\cite{Yang2021DrivingBA}~\cite{Lee2008Detection}~\cite{Kumar2017AVA}, and black spots identification~\cite{ZHANG2019607}. In a summary, discovering the stop spots for the IPPs activities barely been investigated. 

Besides, both the video and audio play a pivotal role ~\cite{Murata2011IIT} to identify abnormality in human behaviors. Compared with GPS, either video or audio directly reflects the human behaviors, \textit{e.g.}, fighting in surveillance scenes~\cite{HA2022}. However, to our best knowledge, how to leverage video or audio to detect the IPPs activity is still an open problem.

\subsection{Handle Low-frequency GPS in Traffic}

Low frequency GPS are application-logged data from GPS embedded equipments on diverse vehicles, \textit{e.g.}, trucks, buses. Different applications enforce diverse priors to handle the information loss caused by the low frequency. For instance, either the connection between the shortest path and the vehicle trajectory~\cite{QUDDUS2015328} or the spatial and the temporal information~\cite{Liu2017GPS}~\cite{LUO20201407} is used as a prior to tackle the low-frequency GPS for the map matching problem. As a comparison, we assume that a vehicle would follow a linearly changed velocity as a prior to discover the possible stop behavior.   

\subsection{Low Rank for traffic problem}

Although transportation in a city is a mixture of the multiple factors, (\textit{e.g.}, the influence of the road conditions, traffic conditions, and climate and environment), recent research has discovered that the spatio-temporal data from the transportation problem has the low-rank structure~\cite{Pangjunbiao2018IEEE}~\cite{Zhe2013UTE}. Consequently,~\cite{CHEN2020102673}~\cite{Yu2021IITS} proposed a non-convex low-rank model to address missing data problem. Non-negative matrix (and tensor) factorization has been widely used for network traffic analysis~\cite{Ahmadi2015ICS}.~\cite{Karve2021SeasonalDI} proposed a method based on sparse non-negative matrix factorization to identify the periodic patterns in traffic behavior. In this paper, we strip the abnormal stop spots from the abnormal ones by the low rank and the sparse decomposition.  

\section{Methods}\label{sec:Methodology}

\subsection{Background}\label{sec:background-problem}

The dataset comprising coach GPS records were sourced from the Beijing transportation information center. 
Specifically, each coach is equipped with a GPS enabled device that transmits real-time but low frequent data including longitude, latitude, time stamps, instantaneous velocity, and engine state (see Table~\ref{tab1:Bus GPS data description}). All items are relayed to a data center via telecommunication networks. 

Adherence to local coach regulations is mandatory, requiring: 1) connectivity of GPS signals; and 2) accurate GPS time stamps. Non-compliance, indicated by erroneous data transmission, prompts law enforcement to notify coach companies for corrective actions. 
  
\begin{mydef}[A GPS point]
A GPS point is a pair of longitude and latitude coordinates with a timestamp collected by an embedded GPS device. In this paper, we additionally collect the instantaneous velocity and the engine state of a coach for each GPS point. Therefore, each GPS point $\bm{p}_{i}$ consists of longitude $\bm{p}_{i}.lng$ , latitude $\bm{p}_{i}.lat$ , timestamp $\bm{p}_{i} .t$ , instantaneous velocity $\bm{p}_{i} .v$ and the engine state $\bm{p}_{i}.of$, as in the example shown in Table~\ref{tab1:Bus GPS data description}.
\end{mydef}

\begin{mydef}[A journey of a coach]
A journey of a coach $\mathcal{T}$ consists of a series of continuous GPS points, and each time interval between any two adjacent GPS points does not exceed a certain value $\Delta t$, which represents the sampling interval; \textit{i.e.}, $\mathcal{T} : \bm{p}_1 \rightarrow \bm{p}_2 \rightarrow \ldots \rightarrow \bm{p}_N $, and $0 < \bm{p}_{i +1} .t - \bm{p}_i .t \le \Delta t (1 \le i \le N )$.
\end{mydef}

\begin{table}[t!] 
\centering
\caption{The items in a GPS point from a long-range coach} \label{tab1:Bus GPS data description}
\begin{tabular}{|c |c|} 
  \hline
  Item  & Meaning \\
  \hline \hline
  Time stamp $t$   & Sending time of a GPS point\\\hline
  Longitude $lng$  & Longitude of a GPS point \\\hline
  Latitude  $lat$ &  Latitude of a GPS point \\\hline
  Instantaneous velocity $v$ & Velocity \\\hline 
  Engine state $of$   & start; off  \\\hline
\end{tabular}
\end{table}

\subsection{Extract candidate stop spots from low-frequency GPS}\label{sec:sub:extract-candidate-stop-spot}

\begin{figure}[t!]
\centering
    \subfloat[case 1:$v_{m}=0$]{\includegraphics[width=.17\textwidth]{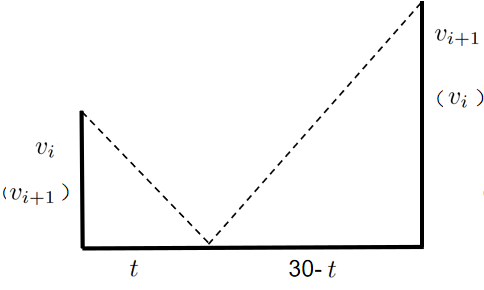}}
    \    \
    \subfloat[case 2: $v_{m}<0$]{\includegraphics[width=.17\textwidth]{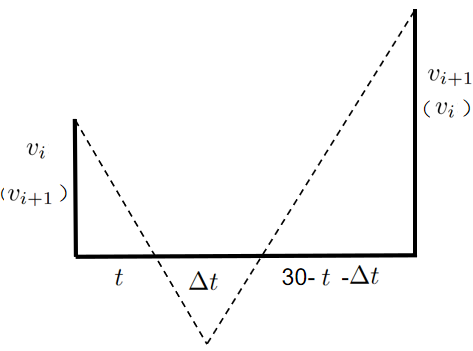}}
  \caption{The illustration of two cases to judge the stop behavior of a coach.}
  \label{fig:Stoping point relation} 
\end{figure}

\begin{mydef}[the low-frequency GPS]
Low-Frequency (or low-sampling-rate) GPS (LF GPS) data fetch a location every 30 seconds or even more. 
\end{mydef}
Obviously, LF GPS makes the location uncertainty increase as the sampling rate reduces. Consequently, judging whether a coach has stopped or not from LF GPS is very difficult due to the long time intervals. If a coach is assumed to linearly change speed, we have two cases which indicate that a coach has been stopped in 30 seconds as in Fig.~\ref{fig:Stoping point relation}.

Concretely, given two consecutive GPS points with the velocity $v_i$ and $v_{i+1}$, the time interval $30(\text{seconds})$ between the consecutive GPS points, and $v_m$ is the minimum velocity between the two consecutive GPS points, a coach would have a stop behavior as follows:

\textbf{Case 1:} when both the minimal velocity $v_m$ is equal to 0 and the stop duration $\Delta t =0$, we have:
\begin{equation}\label{equ2:d and t}
\begin{split}
\frac{(v_i-v_m)\cdot t}{2} + \frac{(v_{i+1}-v_m)\cdot (30-t)}{2}+30\cdot v_m = d,
\end{split}
\end{equation}
where $t$ is the time a coach reaches to the minimum speed $v_m$, and $d$ the distance between the two GPS points (see Fig.~\ref{fig:Stoping point relation}(a)). If the minimal velocity in Eq.~\eqref{equ2:d and t}, $v_m$ is equal to 0, (\textit{i.e.}, $v_m = 0$), we have the following equation:
\begin{equation}\label{equ4:v and d}
\begin{split}
    t\cdot (v_{i}-v_{i+1})= 2d-30 v_{i+1}.  
\end{split}
\end{equation}
Therefore, by applying the constrains $t>0$, we have the following inequality as follows:
\begin{equation}\label{eqt:case-1-inequaltity}
    \frac{2d-30 v_{i+1}}{v_{i}-v_{i+1}}>0.
\end{equation}

\textbf{Case 2:} when the minimal velocity $v_m=0$ and the stop duration $\Delta t > 0$, we could assume that a coach has stopped at the time $t$ as follows (see Fig.~\ref{fig:Stoping point relation} (b), (c) ):
\begin{equation}\label{equ3:m and d}
    \frac{v_i\cdot t}{2}+\frac{v_{i+1}\cdot (30-\Delta t-t)}{2}=d,
\end{equation}
    \begin{equation}
      t\leq   \frac{2d-30v_{i+1}-30v_m}{v_i-v_{i+1}}\  \ \text{where}\   \ v_m <0.
    \end{equation}

The key problem to judging whether a coach has stopped or not is: ﬁnd the conditions that satisfy the non-negativity, \textit{i.e.}, $t>0$ and $\Delta t >0$.

Therefore,~\eqref{equ3:m and d} is re-written as follows:
\begin{equation}\label{equ5:stop time}
    \Delta t=\frac{v_i \cdot t}{v_{i+1}} -t+30-\frac{2 d}{v_{i+1}}.
\end{equation}
To remove the variable $t$, we have the following sub-cases:

\textbf{Sub-case 2.1: $v_{i+1} > v_{i}$:} Given $v_{i+1} > v_{i}$, we have $\frac{v_i}{v_{i+1}}<1$.Consequently,~\eqref{equ5:stop time} is re-written as follows:
\begin{equation}\label{equ7:Inequality stop time}
\begin{split}
    \Delta t &\geq \big(\frac{v_i}{v_{i+1}}-1 \big)\cdot\frac{2d-30v_{i+1}-30v_m}{v_i-v_{i+1}}+30-\frac{2 d}{v_{i+1}}\\
    & \geq \big(\frac{v_i}{v_{i+1}}-1 \big)\cdot\frac{2d-30v_{i+1}}{v_i-v_{i+1}}+30-\frac{2 d}{v_{i+1}}
\end{split}
\end{equation}
Eq.~\eqref{equ7:Inequality stop time} give the low bound of the stop duration $\Delta t$. That is, if the low bound in Eq.~\eqref{equ7:Inequality stop time} is greater than 0, a stop has occurred between the two GPS points.

\textbf{Sub-case 2.2: $v_{i+1} \leq v_i$:} Given $v_{i+1} \leq v_i$, we have $\frac{v_i\times t}{v_{i+1}}-t\geq 0$. Consequently, we simplify drop the term $\frac{v_i\times t}{v_{i+1}}-t$ in ~\eqref{equ3:m and d} as follows:
\begin{equation}\label{equ8:Inequality stop time}
    \Delta t\geq -\frac{2\times d}{v_i}+30.
\end{equation}

\begin{algorithm}[t!]
\small
\caption{Extract Stop duration from the low-frequency GPS}\label{alg:extract_stop}
\begin{algorithmic}[1]
\STATE  \textbf{INPUT:} $v_i,v_{i+1}$, and $d$
\STATE  \textbf{OUTPUT:} Stop signal (Yes/No), and stop duration $\Delta t$ (if stopped)
\IF{~\eqref{eqt:case-1-inequaltity} is hold} 
    \STATE $\Delta t \leftarrow 0$ 
    \STATE \textbf{return} "Yes", $\Delta t$    
\ELSE
    \IF{ $v_{i+1} > v_i$}
    \STATE $\Delta t$ is determined by the Eq.(\ref{equ7:Inequality stop time})
             \IF{$\Delta t > 0$}
             \STATE \textbf{return} "Yes", $\Delta t$
             \ENDIF
    \ELSIF{ $v_{i+1} \leq v_i$}
    \STATE $\Delta t$ is determined by the Eq.(\ref{equ8:Inequality stop time})
            \IF{$\Delta t > 0$}
            \STATE \textbf{return} "Yes", $\Delta t$
            \ENDIF
    \ENDIF
\ENDIF
\STATE \textbf{return} "No"
\end{algorithmic}
\end{algorithm}

\subsection{Group stop spots into a matrix}

Ideally, we expect to precisely discover where a IPP activity occurs. Based on the first Markov assumption in traffic~\cite{Lawlor2017Sensors}~\cite{XU2024PTC}, we adaptively segment the route into fine-grained road segments which help law enforcers precisely locate the IPP spots.

\begin{mydef}[Road segment]
A road segment $r$ is an un-directed edge. Each road segment is associated with a unique ID $r_{id}$, a starting point $\mathbf{r}_{s}$, and an ending point $\mathbf{r}_{e}$. Each point consists of the longitude and latitude coordinates.  
\end{mydef}

In this paper, we uniformly segment a travel line of a coach into the road segments the fixed length $D$. The estimated stop time $\Delta t$ in Alg.~\ref{alg:extract_stop} were assigned to the corresponding road segments. Concretely, if there is are $N$ stop time $\Delta t_{i,s}$, $(s=1,2,...,S)$ of a coach in the $i$-th road segment during one day, we summarize the stop duration of a coach in the $i$-th segment during one day as follows: 
\begin{equation}\label{equ9:Parking matrix}
\bm{R}_{i,j}= \sum_{s=1}^{S}\Delta t_{i,s},
\end{equation}
where $\bm{R}\in \mathbb{R}^{M\times N}$ is the stop duration matrix, $M$ is the number of the road segments, $i=1,2,...,M$. $N$ is the total number of stop durations of each bus in each day after combining the stop durations of the same coach on the same day according to Eq.~\eqref{equ9:Parking matrix}, $j=1,2,...,N$.  

\begin{figure}[t!]
  \centering
    \includegraphics[width=.3\textwidth]{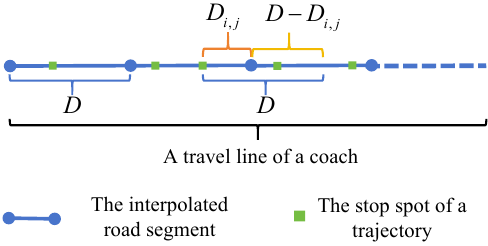}
  \caption{The relationship between the travel line, road segments, stop duration and its interpolated durations.}
  \label{fig:road} 
\end{figure}

\textbf{Dealing with location inaccuracy:}  In many cases, the location information may be miss-located due to the blocked signal transmission. Therefore, a smooth mechanism is preferred over the stop duration $\Delta t_{i,j}$ between the $i$-th interval and $t+1$-th one.
As shown in Fig.\ref{fig:road}, a simple and yet reasonable approach uses linear interpolation to smooth the stop duration $\Delta t_{i,j}$ as follows:
\begin{equation}
\Delta t_{i,j} \leftarrow \frac{D_{i,j}}{D}\times \Delta t_{i,j}, \  \ \Delta t_{i+1,j} \leftarrow \frac{D-D_{i,j}}{D}\times \Delta t_{i,j},
\end{equation}
where $D_{i,j}$ is the distance from the stop spot of a trajectory to the end point of the $i$-th road segment.

\subsection{Identify the IPPs spots by the low rank constraint} 

In order to strip the Abnormal Stop Spots (ASSs) from the matrix $\bm{R}$, we introduce the stop type matrix $\bm{I}\in \mathbb{R}^{M\times N}$, where $\bm{I}_{ij}\in \{0,1\}$, in which $0$ means the abnormal stop, and $1$ indicates the normal stop. Consequently,
striping the abnormal stops from the normal ones is modeled as follows:
\begin{equation}\label{equ12:low rank model}
\begin{split}
    \arg \min_{\bm{I},\bm{E}} \quad &\|\bm{\Theta}\|_{*} +\lambda \| \bm{E}\|_{1} + \beta \|\bm{W}\|_{2,1} \\
s.t.: \quad &\bm{R} = \bm{R}\odot \bm{I} + \bm{E}, \\
     & \bm{\Theta} = \bm{R}\odot \bm{I}, \\
     & \bm{E} = \bm{W}, \\
     & 0\leq \bm{E}, 0\leq \bm{I}_{ij}\leq  1,
\end{split}
\end{equation}
where the operation $\odot$ is the element-wise multiplication between two matrices, $\|\cdot\|_1 $ is the $L_1$ norm, $\|\cdot\|_{2,1}$ is the group sparsity, $\|\cdot\|_{*}$ is the nuclear norm, $\lambda (\lambda >0)$ and $\beta (\beta >0)$ are the hyper-parameters. In Eq.~\eqref{equ12:low rank model}, we relaxes the 0-1 programming of $\bm{I}$.

Eq.~\eqref{equ12:low rank model} assumes that the stop duration matrix $\bm{R}$ is decomposed into a low rank matrix $\bm{\Theta}=\bm{R}\odot \bm{I}$, and a mixture of the sparse matrix $\bm{E}$ and column-sparse matrix $\bm{W}$. Motivated by~\cite{Pangjunbiao2018IEEE}, the matrix $\bm{\Theta}$ in Eq.~\eqref{equ12:low rank model} represents the normal stop spots, and the sparse matrix $\bm{E}$ indicates the abnormal stop spots. Therefore, the column-sparse $\|\cdot\|_{2,1}$ models that some road segments are impossible to be ASSs, while the sparse $\|\cdot\|_1$ follows that the ASSs should be sparse compared to the normal stop behaviors of coaches. In summary, Eq.~\eqref{equ12:low rank model} assumes that the ASSs are sparely occurred in some special road segments. Therefore, a larger value of $\bm{E}_{ij}$ indicates a higher ASS for the $i$-th road segment.

\subsection{Model Optimization}\label{sec:optimization}

We use the Alternating Direction Method of Multipliers (ADMM) to solve Eq.~\eqref{equ12:low rank model} as follows:
\begin{equation}\label{equ13:Lagrange function}
\begin{split}
L(\bm{I},\bm{E},\bm{W}) =& \| \bm{\Theta} \|_{*} + \lambda \| \bm{E}\|_{1} +
+ \beta \|\bm{W}\|_{2,1} \\
&+ \langle \bm{Y}_{1}, \bm{R}-\bm{R}\odot \bm{I} - \bm{E} \rangle \\
&+  \langle \bm{Y}_{2}, \bm{\Theta}-\bm{R}\odot \bm{I}  \rangle + \langle \bm{Y}_{3}, \bm{W}-\bm{E} \rangle \\
&+ \frac{\rho}{2}\|\bm{R}-\bm{R}\odot \bm{I} - \bm{E}\|_{F}^{2} \\
&+\frac{\rho}{2}\| \bm{\Theta}-\bm{R}\odot \bm{I}  \|_{F}^{2} +\frac{\rho}{2}\| \bm{W}-\bm{E} \|_{F}^{2},
\end{split}
\end{equation}
where $\bm{Y}_1\in \mathbb{R}^{M\times N}$, $\bm{Y}_2\in \mathbb{R}^{M\times N}$, and $\bm{Y}_3\in \mathbb{R}^{M\times N}$ are the Lagrange multiplier matrices, and $\rho$ ($\rho >0$) is the penalty factor. The iterative alternating direction method is used to solve the Lagrangian function in Eq.~\eqref{equ13:Lagrange function}.

\textbf{Update the matrix $\bm{E}$:} When other variables are fixed, the subproblem \textit{w.r.t.} $\bm{E}$ is as follows:
\begin{equation}~\label{equ14:optimize e}
\arg \min_{\bm{E}}\frac{\lambda}{2} \| \bm{E}\|_{1} + \frac{\rho}{2} \big\|\bm{E}-\frac{1}{2}(\bm{R}+\bm{W}+\frac{\bm{Y}_1^{\top}}{\rho} 
+\frac{{\bm{Y}_3}^{\top}}{\rho}-\bm{R}\odot \bm{I})\big\|_{F}^{2}, 
\end{equation}
which can be solved by the shrinkage method as follows:
\begin{equation}\label{equ17:}
\bm{E} = S_{\lambda / 2\rho}\left(\frac{1}{2}(\bm{R} + \bm{W} +\frac{\bm{Y}_{1}}{\rho}+\frac{\bm{Y}_{3}}{\rho}-\bm{R}\odot \bm{I})\right),
\end{equation}
in which the shrinkage factor ${S}_{\alpha}(Z)$ is defined as follows:
\begin{equation}\label{equ18:}
S_\alpha(x) = \left\{ \begin{array}{ll}
x - \alpha \quad\quad\quad & x >\alpha \\
0  & |x| \leq \alpha \\
x + \alpha & x < -\alpha
\end{array} \right.
\end{equation}

\textbf{Update the matrix $\bm{\Theta}$:} When other variables are fixed, the subproblem \textit{w.r.t.} $\bm{\Theta}$ is as follows: 
\begin{equation}\label{equ19:Opt}
 \arg \min_{\bm{\Theta}} { \| \bm{\Theta} \|}_{\ast }+\frac{\rho }{2} \left\| \bm{\Theta}-\bm{R}\odot \bm{I}
+\frac{{\bm{Y}}_{2}}{\rho } \right\|_{F}^{2},
\end{equation}
which can be solved by the singular value threshold
method. More specially, let $\bm{C}= \bm{U} \bm{\Sigma}\bm{V}^\top $ be the Singular Value Decomposition (SVD) of $\bm{C}$. Then, for ${T_\lambda }(\bm{C}) = \bm{U}{\bm{\Sigma}_\lambda }{\bm{V}^T}$, we have:
\begin{equation}\label{equ19:Opttha}
{T_\lambda }(\bm{C}) = \arg \mathop {\min }\limits_{\bm{\Theta}}  \frac{1}{2}\left\| {\bm{\Theta}  - \bm{C}} \right\|_F^2 + \lambda {\left\| \bm{\Theta}  \right\|_*},
\end{equation}
where ${\bm{\Sigma}_\lambda }$ is diagonal with ${({\bm{\Sigma} _\lambda })_{ii}} = \max (0,{\bm{\Sigma} _{ii}} - \lambda )$. By the SVD threshold method~\cite{dong2012nonlocal}, the solution for the $\bm{\Theta}$ is as fallows:
 \begin{equation}\label{equ21:}
 \begin{split}
    \bm{\Theta}& =\arg \min_{\bm{\Theta}}{L(\bm{\Theta})} \\ 
     &={T}_{\frac{1}{\rho }} \big(\bm{R}\odot \bm{I}-\frac{{\bm{Y}}_{2}}{\rho }\big) . 
 \end{split}
 \end{equation}

\textbf{Update the matrix $\bm{I}$:} When other variables are fixed, the subproblem \textit{w.r.t.} $\bm{I}$ is as follows:

\begin{equation}\label{equ22:}
\begin{split}
    \arg \min_{\bm{I}}& \frac{\rho }{2}({(\bm{R} \odot \bm{I})}^{2}-2{\bm{R}}^{\top}\bm{R}\odot \bm{I}-2{\bm{E}}^{\top}\bm{R}\odot \bm{I}) \\
    &+\frac{\rho }{2}({(\bm{R}\odot \bm{I})}^{2}-2\bm{\Theta}^{\top} \bm{R}\odot \bm{I})-{\bm{Y}}_{1}^{\top}\bm{R}\odot \bm{I} \\
    &-{\bm{Y}}_{2}^{\top}\bm{R}\odot \bm{I},
\end{split}
\end{equation}
Then $\bm{I}$ is solved as follows:
\begin{equation}\label{equ23:}
    \begin{split}
        \bm{I} = ({\rho \bm{R} + \rho \bm{\Theta} - \rho \bm{E} + \bm{Y_1} + \bm{Y_2}})/({2\rho \bm{R}}),
    \end{split}
\end{equation}
where the operation $/$ represents the element-wise division between two matrices.

\begin{algorithm}[t!]
\small
\caption{Optimize our model with ADMM}\label{alg:optimize_stop}
\begin{algorithmic}[1]
\STATE \textbf{INPUT:} The matrix$\bm{R}$, $\lambda > 0$, $\beta > 0$ and $\rho \geq 1$;
\WHILE{not converged}
    \STATE Update the matrix $\bm{E}$ by Eq.(\ref{equ17:});
    \STATE Update the matrix $\bm{\Theta}$ by Eq.(\ref{equ21:});
    \STATE Update the matrix $\bm{I}$ by Eq.(\ref{equ23:});  
    \STATE Update the matrix $\bm{W}$ by Eq.(\ref{equ26:});
   \STATE Update the Lagrange multiplier matrices $\bm{Y}_1$, $\bm{Y}_2$, and $\bm{Y}_3$ by Eq.~\ref{equ27:};
   \STATE $\rho \leftarrow \rho\cdot \mu$;
\ENDWHILE
\STATE \textbf{OUTPUT:} $\bm{I}, \bm{E} $ 
\end{algorithmic}
\end{algorithm}

\textbf{Update the matrix $\bm{W}$:} When other variables are fixed, the subproblem \textit{w.r.t.} $\bm{W}$ is as follows:
\begin{equation}\label{equ24:}
\begin{split}
     \arg \min_{\bm{W}} \beta \| \bm{W} \|_{2,1}+\frac{\rho }{2} \left\| \bm{W}-\bm{E}+\frac{{\bm{Y}}_{3}}{\rho }\right \|_{F}^{2}.
\end{split}
\end{equation}
where $\| \bm{W} \|_{2,1} = \sum\limits_{i = 1}^n {\sqrt {\sum\limits_{j = 1}^m {\bm{W}_{ij}^2} } }  = \sum\limits_{i = 1}^n {{{\left\| {{\bm{W}_i}} \right\|}_2}}$. We can update the $i$-th column of the $\bm{W}$ matrix as follows:
\begin{equation}\label{equ26:}
{\bm{W}}(:,i) = \left\{ \begin{array}{ll}
\frac{ \|{\mathbf{Q}}_{i}\|-\alpha }{\|\mathbf{Q}_{i}\|}\mathbf{Q}_{i},  & \
if  \  \ \alpha < \|\mathbf{Q}_{i}\|,   \\
0,      & \text{otherwise},   
\end{array} \right.
\end{equation}
where $\alpha =\beta/\rho$, $\mathbf{Q}=\bm{E}-\frac{{\bm{Y}}_{3}}{\rho }$, $\mathbf{Q}_i$ is the $i$-th column of the matrix $\bm{Q}$.

\textbf{Update the matrix $\bm{Y}$:} 
Update the Lagrange multipliers ${\bm{Y}}_{1}$, ${\bm{Y}}_{2}$ and ${\bm{Y}}_{3}$ are as follows:
\begin{equation}\label{equ27:}
\begin{split}
     &{\bm{Y}}_{1}\leftarrow {\bm{Y}}_{1}+\rho(\bm{R}-\bm{R}\odot \bm{I}-\bm{E}),  \\
    &{\bm{Y}}_{2}\leftarrow{\bm{Y}}_{2}+\rho(\bm{\Theta} -\bm{R}\odot \bm{I}),  \\ 
       &{\bm{Y}}_{3}\leftarrow{\bm{Y}}_{3}+\rho(\bm{W}-\bm{E}),
\end{split}
\end{equation}

During training, we update $\rho \leftarrow \rho\cdot \mu$, where $\mu > 1$. In this paper, we initialized $\rho$ to 1 and set $\mu$ to 1.2. The ADMM process is summarized in Alg.~\ref{alg:optimize_stop}.

\section{Experiments and Case study}\label{sec:Exp and dis}

\subsection{Experiment setup}

\subsubsection{Data} 
\begin{figure}[t!]
  \centering
    \includegraphics[width=.48\textwidth]{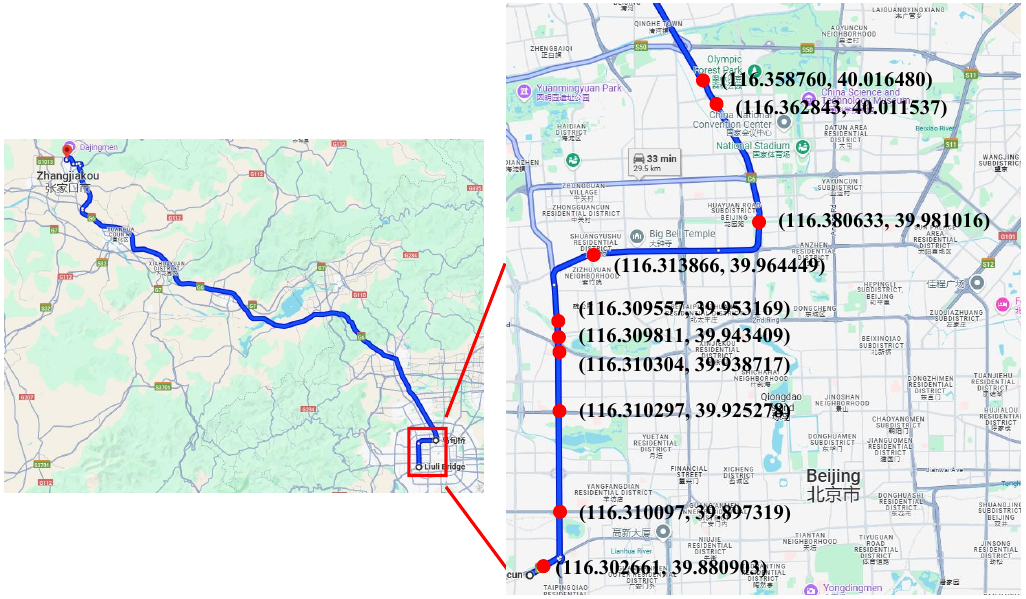}
  \caption{ Visualization of the manually discovered ASSs.}
  \label{fig:Actual abnormal stop point location} 
\end{figure}

We use the GPS data generated from the travel line with the Liuliqiao start station in Beijing to Zhangjiakou end station in Hebei province for our study. Since the coaches barely stop on highway, we have invited 11 students to take this travel line where the coaches have the different departure times. These students have recorded the ASSs for the IPPs activities. The GPS points of these abnormal stops are recorded by the APP in mobile phones. Fig.~\ref{fig:Actual abnormal stop point location} shows that 10 recorded abnormal stop spots. 

During the pre-processing, we used three conventional tricks: 1) the area constraint, 2) the instantaneous speed, and 3) the precision of GPS. Concretely, the area constraint assumes that GPS only occurs within the Sixth Ring Road of Beijing. The instantaneous speed means that we should remove the GPS points who's instantaneous speed is larger than 120 $km/h$. The precision of GPS means that we remove error GPS data if the number of digits after the point is less than 6.  

\subsubsection{Experiment settings} 
 In our experiments, our experiments are conducted on a server (Windows) with Intel Core CPU with 3.2 GHz and 32 GB main memory. The source code was implemented with Matlab and Python. 

\subsubsection{Evaluation metrics} In our experiments, we use two kinds of evaluation metrics as follows:
\begin{itemize}
\item \emph{Average Precision (AP):} We plot Precision versus Recall of a Classifier (PRC) at different decision thresholds. Consequently, AP calculates the average precision across different decision thresholds, achieved by calculating the area under the PRC. higher AP is, better a method is.

\item \emph{Area Under Curve (AUC):} Receiver Operating Characteristic (ROC) curve plots True Positive Rate (TPR) versus False Positive Rate (FPR) of a classifier at different decision thresholds. Consequently, the AUC value represents the area under the ROC curve. the larger the area under the curve is, the closer the AUC value is to 1.
\end{itemize}

\subsection{Baselines and the State-of-Art Approaches}\label{sec:sub:baseline}

\subsubsection{Exploring how to leverage the sparse matrix}\label{exp_matrix} Three indicators respectively are proposed to judge whether an abnormal stop occurs or not as follows:
\begin{itemize}  
\item  \textit{All Stop Time (AST):} The sum of the abnormal stop durations in the matrix $\bm{E}$, \textit{i.e.},  $\sum_{j}\bm{E}_{i,j}$ for the $i$-th road segment. AST assumes that different coach drivers would spontaneously select the same spot for the IPPs activity.
\item \textit{Maximum Stop Time (MST):} The max stop duration time of a coach in the matrix $\bm{E}$, \textit{i.e.}, $\max_{j}\bm{E}_{i,j}$ for the $i$-th road segment. MST assumes that the largest stop duration would be the ideal abnormal stop spots for the IPPs activity.
\item \textit{Top-$k$ Averaged stop Time (TAT@$k$):} To remove the possible noisy in the matrix $\bm{E}$, TAT@$k$ selects the top-$k$ stop duration at the $i$-th road segment, \textit{i.e.}, $\frac{1}{k}\sum_{i=1}^k topk(\bm{E}_{i,:})$, where the function $topk(\bm{E}_{i,:})$ finds the top-$k$ values from the $i$-th column of the matrix $\bm{E}$. TAT@$k$ assumes that a few stops are caused by the non-abnormal stop reason. Therefore, TAT@$k$ is the a version of AST with the noise remove ability. 
\end{itemize}

\subsubsection{The possible SOTA solutions}\label{sota_solution} To our best knowledge, we found that the off-the-shelf methods barely be directly used to our problem. In fact, stripping abnormal stops from a matrix is mathematically equal to matrix decomposition. Therefore, this paper proposed the three possible SOTA solutions and their reasons as follows:
\begin{itemize}
\item \textit{Without Structure Sparsity (WST):} WST ignores the physical constraints of spots as follows:
\begin{equation}\label{equ30:wst}
\begin{split} 
  \arg\min_{\bm{I},\bm{E}}\quad & {rank(\bm{\Theta} )}+\lambda \|\bm{E}\|_{1} 
  \\s.t.&\bm{R} =\bm{R}\odot \bm{I}+\bm{E}, 
  \\&\bm{\Theta} =\bm{R}\odot \bm{I},  
  \\ &0 \le \bm{E},0\leq \bm{I}_{ij}\leq 1.
 \end{split}
\end{equation}
WST assumes that the abnormal stops could occur at any spot. The comparisons between WSA and our method can reveal the importance of the physical constraints motivated group sparsity. 

\item \textit{Without Stripping Abnormal Stop (WSA):} The stop duration matrix $\bm{R}$ is directly used to discover the abnormal stop spots. The WSA strategy is widely used in the black spots
identification task~\cite{FAN2019607}. WSA does not assume that the IPPs activity is sparsely occurring. The comparisons between WSA and our method reveal whether ASSs are sparse, and whether matrix factorization could separate ASSs from the normal stop behaviors. 

\item \textit{Using Instantaneous Speed of GPS (UIS):} When the instantaneous speed of a coach is equal to 0, a stop spot is recorded by a GPS point. Concretely, the corresponding stop duration matrix $\bm{R}$ is firstly built by the zero instantaneous speed GPS. Second, we utilized the proposed method in Eq.~\eqref{equ12:low rank model} to discover the ASSs. The difference between UIS and our method is whether handing the low-frequency GPS or not. The comparisons between UIS and our method can reveal that the model that we extract the stop candidates from low-frequency GPS data are reasonable.
\end{itemize}

For a fair comparison, the other hyper-parameters of these SOTA methods are tuned to obtain the best performances.

\subsection{Analysis of our approach}\label{sec:sub:analysisAR3}

\subsubsection{Selection of the abnormal stop indicators}

We compared three indicators in Section~\ref{exp_matrix} when the proposed model in Eq.~\eqref{equ12:low rank model} has used with the a randomly guessed hyper-parameter, \textit{e.g.}, the sparsity weight $\lambda =0.3$, and the group sparsity $\beta =0.3$. Table \ref{tab:ap_auc} details the performances of three indicators at the different length of the road segments, uncovering the 2 observations as follows: 

\begin{table}[t!]
\centering
\caption{The AP and AUC values for the different size (meters) of road segments.} 
\label{tab:ap_auc}
\resizebox{\linewidth}{!}{
\begin{tabular}{c|ll|ll|ll|ll}
\hline
Size & \multicolumn{2}{c|}{100}                          & \multicolumn{2}{c|}{200}                          & \multicolumn{2}{c|}{300}                          & \multicolumn{2}{c}{400}                          \\ \hline
          Indicators & \multicolumn{1}{c}{AP} & \multicolumn{1}{c|}{AUC} & \multicolumn{1}{c}{AP} & \multicolumn{1}{c|}{AUC} & \multicolumn{1}{c}{AP} & \multicolumn{1}{c|}{AUC} & \multicolumn{1}{c}{AP} & \multicolumn{1}{c}{AUC} \\ \hline
AST        & \bf{0.1833}                 & \underline{0.1875}                   & \bf{0.5556}                 & \bf{0.7619}                   & \underline{0.4167}                 & \underline{0.6875}                    & \bf{0.7750}                 & \bf{0.7500}                  \\
MST        & 0.1000                 & 0.0556                   & \underline{0.5317}                 & \underline{0.7143}                   & \bf{0.4500}                 & \bf{0.7500}                   & 0.7222                 & 0.5833                  \\
TAT@2      & \underline{0.1250}                 & \bf{0.2222}                   & \underline{0.5317}                 & \bf{0.7619}                   & \underline{0.4167}                 & \underline{0.6875}                   & \underline{0.7250}                 & \underline{0.6250}                  \\ \hline
\end{tabular}}
\end{table}

\begin{itemize}
    \item \textit{The AST indicator achieves the best performance among these indicators in both AUC and AP in most cases.} The explanation is that different coach drivers tend to consensually choose the same spots where the IPPs activity occurs. Compared to the other indicators, the sum operation in AST could suppress the noises in the abnormal stop situation. 
    
    \item \textit{AST slightly outperform MST and TAT@2.} For example, when the size of the road segment is 200 meters, values of the AST indicator are 4.8\% and 2.4\% higher than that of MST in terms of both AUC and AP, respectively; while AST surpassed TAT@$2$ 0\% and 2.4\% in terms of both AUC and AP, respectively. The comparative results reflect the viewpoint that ``social support among survivors is common in mass emergencies and can be facilitated by action by professional groups''~\cite{DRURY202012}. That is, different coach drivers would gradually choose a few preferable spots where the transport infrastructures and the road conditions facilities the occurrence of the IPPs activity.
\end{itemize}
In this paper, we use the AST as the indicator for the following experiments. 

\begin{figure*}[t!]
  \centering
  \subfloat[200 meters]{ \includegraphics[width=.45\textwidth]{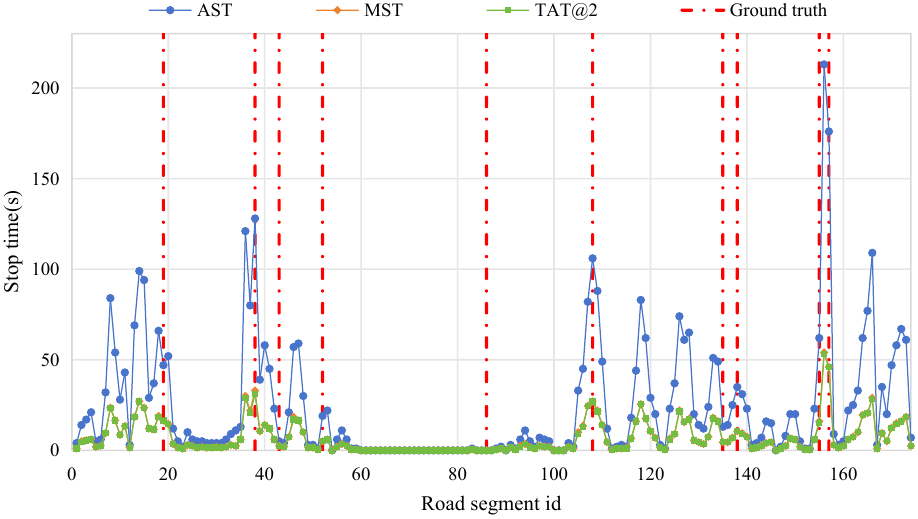}}
  \subfloat[300 meters]{\includegraphics[width=.45\textwidth]{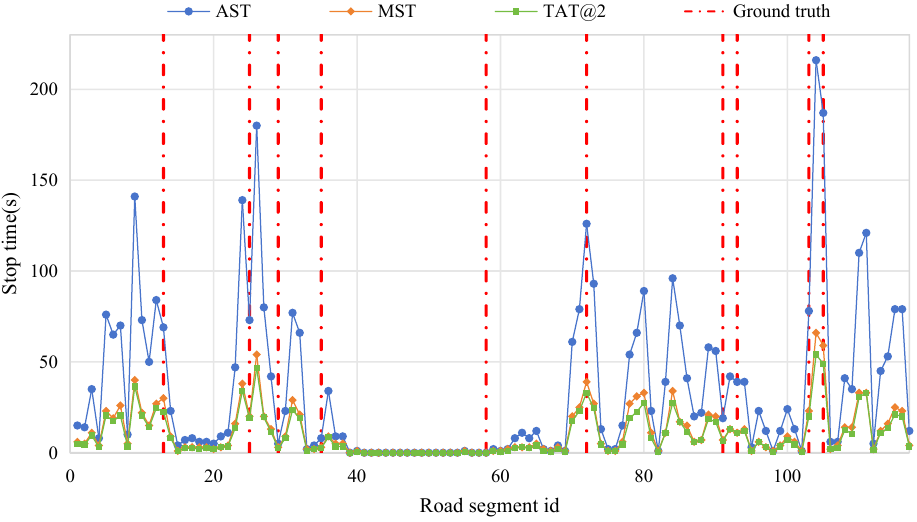}}
  \caption{The values of the three indicators for every road segment when the interval is either 200 or 300 meters.}
  \label{fig:interval} 
\end{figure*}

\subsubsection{Effectiveness of the intervals}\label{sec:eff_interval}

Table.~\ref{tab:ap_auc} also detailed the effectiveness of the different length of road segments. We have observed that the lager size of road segments would bring a better result than that of the smaller one. The main exploitation is that it is easier to predict the ground truth in a larger road segment. However, as the size of road segments increases, identifying the exact locations of the IPP activity becomes more challenging. In this paper, we used the road segments with 200 meters to balance between the accuracy of the model and the difficulty to identify the IPPs spots.

Interestingly, we have observed that the performance slightly dropped when the interval was 300 meters. Fig.~\ref{fig:interval} further shows the detailed values of the three indicators for different size of road segments. When the road segment was 300 meters, the optimized abnormal stop duration at the ground truth road segment is significantly different than that of when the road segment is 200 meters in Fig.~\ref{fig:interval}. The explanation is that the size of road segments would slightly change the distribution of the stop duration matrix, although the location inaccuracy is adopted.

\begin{figure}[h!]
  \centering
\subfloat[The effectiveness of $\lambda$ when $\beta=0.1$]{ \includegraphics[width=.24\textwidth]{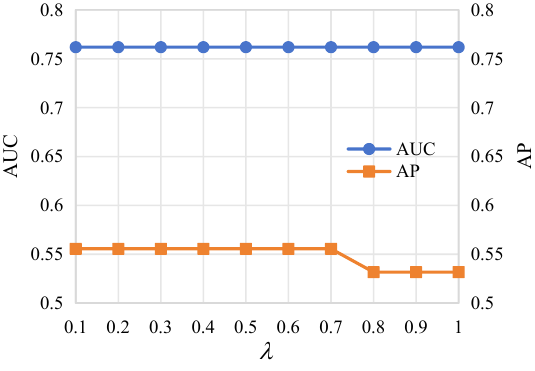}}
  \subfloat[The effectiveness of $\beta$ when $\lambda=0.1$]{  \includegraphics[width=.24\textwidth]{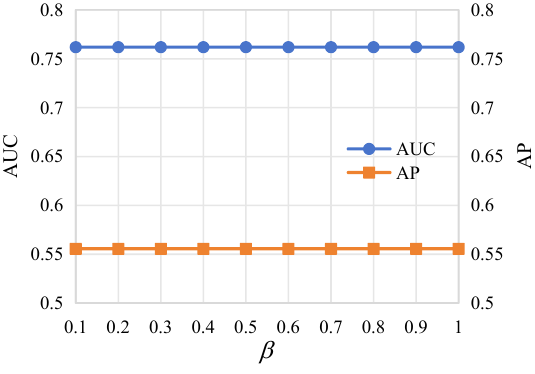}}
  \caption{The effectiveness of the $\beta$ and $\lambda$.}
  \label{fig:alpha-beta-effective} 
\end{figure}

\subsubsection{Influence of $\lambda $ and $\beta $} 

Fig.~\ref{fig:alpha-beta-effective} shows that the impacts of the different settings of $\lambda$ and $\beta$. In Fig.~\ref{fig:alpha-beta-effective}(a), we fixed $\beta=0.1$ and set the $\lambda$ ranging for 0 to 1 with the step size 0.1.
We observed that our method is very stable under the different hyper-parameters.
Similarly, Fig.~\ref{fig:alpha-beta-effective}(b) fixed $\lambda=0.1$ and investigated the impact of $\beta$. 
Fig.~\ref{fig:alpha-beta-effective}(b) shows that both AP and AUC remain unchanged under the different $\beta$ settings. This indicates that the non-negative $\beta$ has little impact on the model.

Moreover, the comparisons between Fig.~\ref{fig:alpha-beta-effective}(a) and~\ref{fig:alpha-beta-effective}(b) discovered that $\lambda$ has a greater impact than $\beta$. For instance, there is a little performance drop in terms of AP when $\lambda$ is larger than 0.8 in Fig.~\ref{fig:alpha-beta-effective}(a), while $\beta$ shows no effect on the performance in the range of [0, 1] in Fig.~\ref{fig:alpha-beta-effective}(b). These results shows the the the sparsity term is more important the group sparsity one in Eq.~\eqref{equ12:low rank model}. The explanation is that some spots are impossible to be selected as the candidate for the IPPs activity. The zero stop duration times for the \#60 to \#80 road segments in Fig.~\ref{fig:interval}(a) also verify this phenomenon. In this paper, we simply used the following hyper-parameters: $\lambda=0.1$ and $\beta=0.1$ in the following experiments.

\subsubsection{Analysis of the convergence speed} 
To reduce the scale of problem, we firstly grouped road segments into several cluster by performing Affinity Propagation (AP) clustering~\cite{frey2007clustering} on GPS points. In our experiments, we have 23 clusters which ranges from 1.5 kilometers (KM) to 6 KM. That is, we can solve Eq.~\eqref{equ12:low rank model} in a distributed parallel strategy.  

\begin{figure}[t]
  \centering
    \includegraphics[width=.3\textwidth]{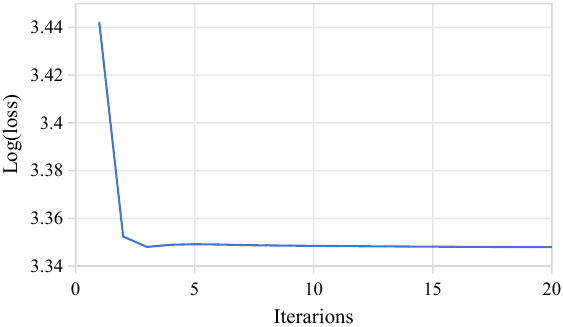}
  \caption{Convergence curves of one cluster}
  \label{fig:The loss} 
\end{figure}

Fig.~\ref{fig:The loss} illustrates the convergence speed of the matrix with $164 \times 24$ in Eq.~\eqref{equ12:low rank model}. Fig.~\ref{fig:The loss} discovers that the optimization method in Alg.~\ref{alg:optimize_stop} would rapidly converge to a local minimum. For instance, the loss values decreased rapidly as the number of iterations increases from 0 to 5 in Fig.~\ref{fig:The loss}, where each iteration costed 0.004 seconds. Although the proposed method leverages ADMM to iteratively solve several sub-problems, the fast solvers for each sub-problem still guarantee the fast convergence speed in Alg.~\ref{fig:The loss}. The results indicate that the propose method would scale up to a large-scale problem in a parallel system.   

\begin{table}[h]
\centering
\caption{Comparisons between the SOTA approach and our method.} 
\label{tab3:Comparation}
\begin{tabular}{c| c c}
  \hline
  Methods & AUC & AP \\
  \hline \hline
  WST & 0.7143 & 0.5317 \\
  WSA & 0.5238 & 0.3694 \\
  UIS & 0.5312 & 0.2678 \\\hline
  Our method & \textbf{0.7619} & \textbf{0.5556}\\
  \hline
\end{tabular}
\end{table}

\subsection{Qualitative Comparisons with the SOTA Methods}

Table~\ref{tab3:Comparation} compared the SOTA method in Section \ref{sota_solution} with our method. Table~\ref{tab3:Comparation} shows that proposed method achieved the best AUC and AP values. The results indicate that our proposed method can effectively improve the accuracy of capturing abnormal stop spots. Besides, the superior performances of our method uncovered three observations as follows:
\begin{itemize}
    \item \textit{Group sparsity is very necessary to the ASS task.} Our method significantly outperforms WST by 4.76\% and 2.39\% in terms of AUC and AP, respectively. This indicates that the group sparsity in Eq.~\eqref{equ12:low rank model} is very important to discover ASSs. The group sparsity empowers the model to remove the impossible stop spots. 
    
    \item \textit{Stripping abnormal stop duration from the stop duration is necessary.} Our method significantly outperforms WSA, for instance, AUC and AP of the WSA are 23.81\% and 18.62\% lower than that of the proposed method, respectively. This means that there are many normal stop duration are mixed in the $\bm{R}$ matrix builded from strategy in WSA. Moreover, the significant performance gap also verifies that the low rank decomposition efficiently strip the abnormal stop duration form these normal ones. 
   \item \textit{Many abnormal stop behaviors barely been discovered by the instantaneous speed.} Our method significantly outperforms UIS, for instance, AUC and AP of UIS are 23.07\% and 28.78\% lower than that of the proposed method, respectively. Because some abnormal stop would happen in 30 seconds. The comparisons between UIS and our method verify the effectiveness of the necessary and effectiveness of extracting candidate stop spots from low-frequency GPS in Section~\ref{sec:sub:extract-candidate-stop-spot}.
   \end{itemize}

\begin{figure}[t!]
  \centering
\subfloat[ \#156 road segment.]{ \includegraphics[width=.35\textwidth]{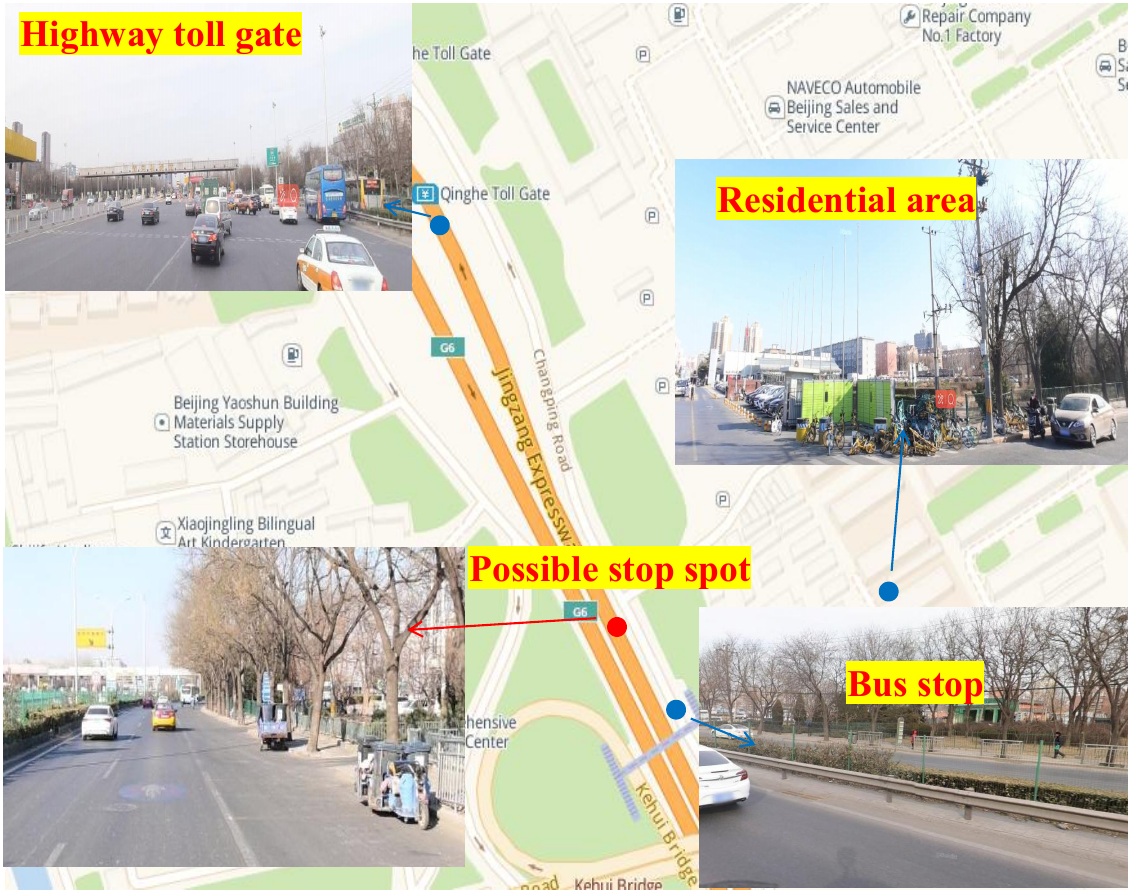}}\\
  \subfloat[ \#157 road segment.]{  \includegraphics[width=.35\textwidth]{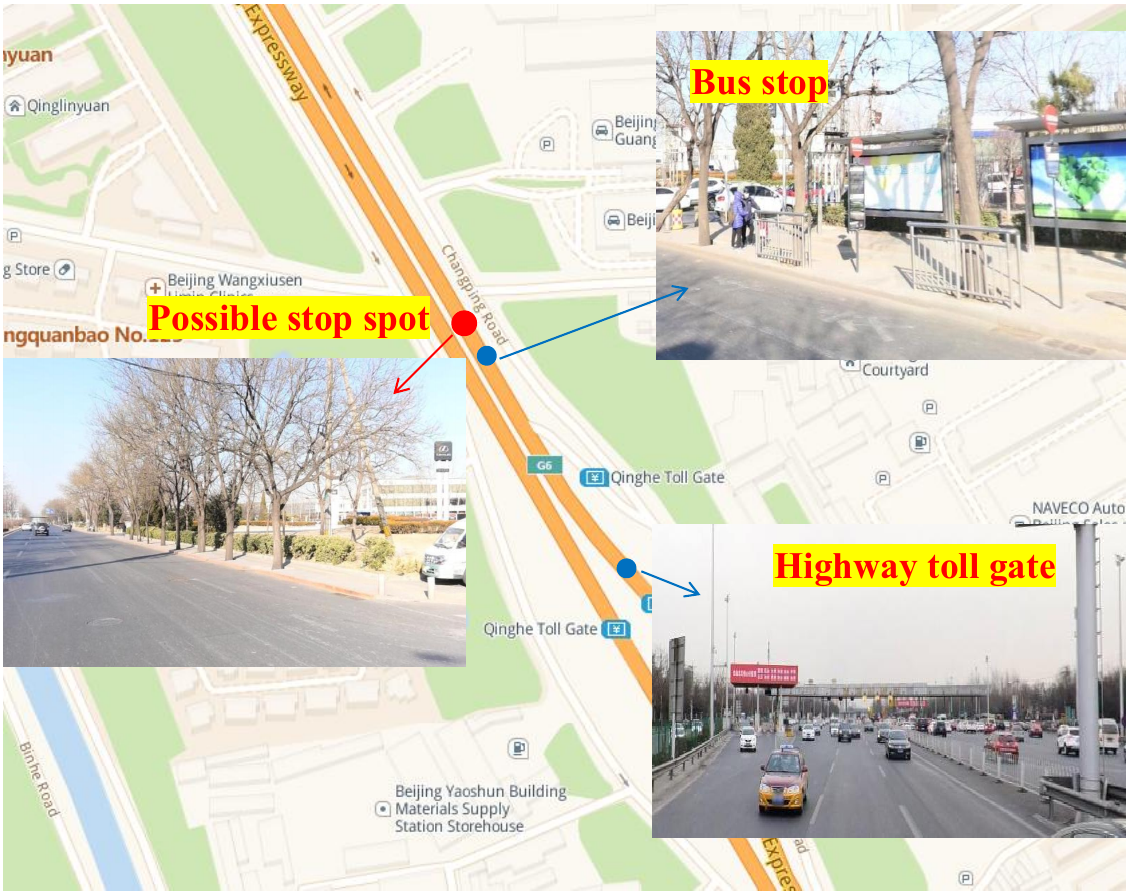}}\\
  \subfloat[\#38 road segment.]{  \includegraphics[width=.35\textwidth]{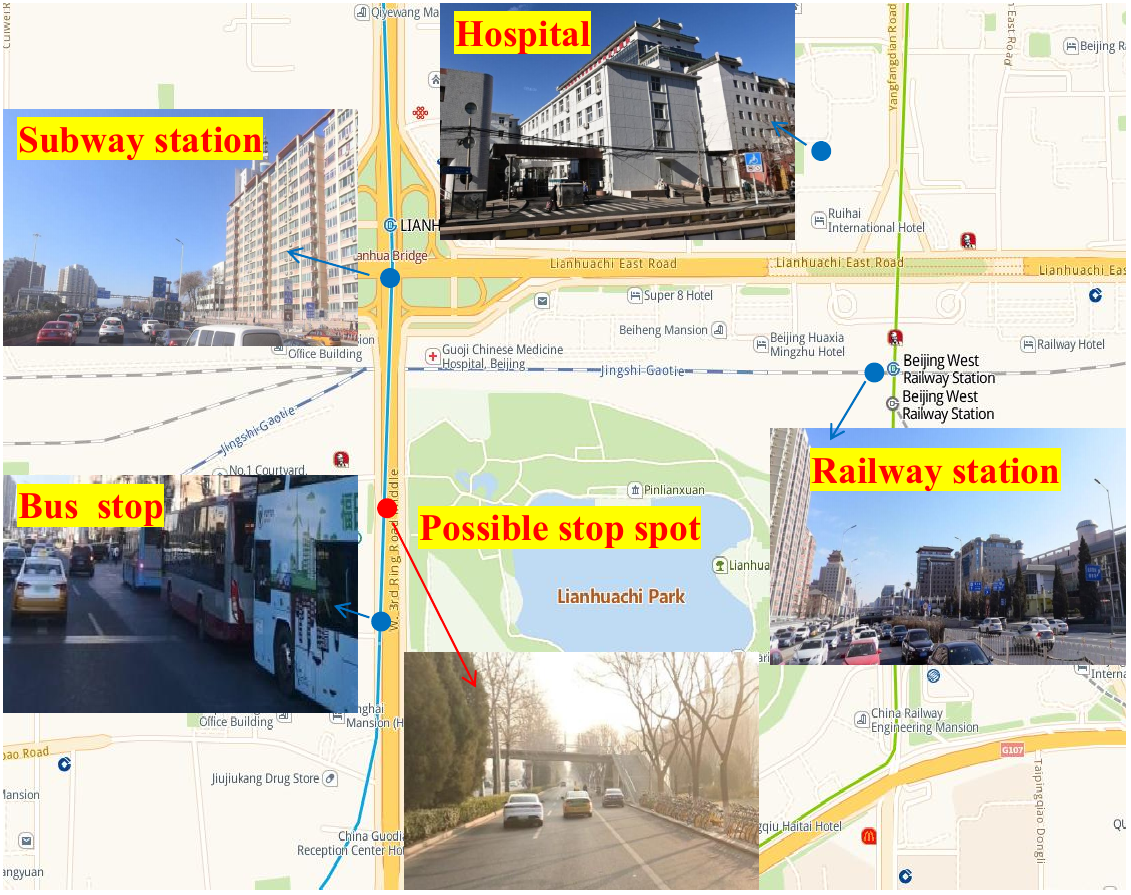}}
  \caption{The street-view of the road segments with the top-3 largest AST value.}
  \label{fig:case}
\end{figure}

\subsection{An case study with discovered spots}

Intuitively, the limited number of law enforcers should be dynamically allocated to the spots where coaches frequently stop for the IPPs activity. In terms of our approach, the AST of the $\bm{E}$ matrix reflects the possibility and the duration of the IPPs activity in a road segment. Except these well-understood spots (\textit{i.e.}, intercity bus stations, railway station, airports, and centralized business areas), the proposed method could effectively discover some spots which have the high probability to occur the IPPs activity. For instance, Fig.~\ref{fig:case} illustrated three street-view of the road segments (\textit{i.e.},\#156, \#157, \#38 in Fig.~\ref{fig:interval}(a)) with the top-3 largest AST value. Note that \#38, \#157 road segments are the ground truth; while \# 156 road segment is the IPPs spot discovered by our method. Specially, Figs.~\ref{fig:case}(a) and~\ref{fig:case}(b) shown that the stop spot is located near a highway toll station, facilitating coaches pick up passengers who have the long range travel demand. Besides, there is a bus stop which makes it convenient for shuffling passengers by buses. Moreover, several intersections provide a space for coaches to pick up passengers. For Fig.~\ref{fig:case}(c), the street-view images show that the stop spot is near an bus stop, subway station and the train station, which help coach drivers to pick up passengers. Because this area has a high probability to pick up the passengers with the long range travel demand. Additionally, there is a long side road which supplies a convenience for coaches to pull over. 


\section{CONCLUSION}\label{sec:conclusion}

In this paper, we have described a method to discover the spots where the IPPs activity tend to occur for long range  coach industry. Concretely, leveraging low-frequency GPS data, the proposed method efficiently identifies the the spots of the IPPs activity in an unsupervised approach by formalizing the challenge problem as the matrix factorization problem. There are significant contributions of the proposed method as follows:
\begin{itemize}
\item To our best knowledge, we firstly introduce the IPPs problem, and treat this challenge as an unsupervised problem. The unsupervised method avoids to harvest the time-costing annotation, making our method practical to leverage the large scale GPS data from long range coaches. 
\item We frame our search for ASSs as the matrix decomposition problem, where the low rank models the normal stops, a sparse matrix captures the abnormal ones, and a group sparsity constrains that some spots would never happen the IPPs activity. 
\item To release the power of the low-frequent GPS, we propose an simple yet efficient method to discover the stop duration. Experimental results verify the effectiveness the linear change of velocity to discover the stop spots.
\end{itemize}

The experimental results demonstrate that the method not only accurately detects locations of the IPPs activity but also exhibits strong robustness. Compared to other methods, our approach has significantly improved detection accuracy. Furthermore, the method proposed in this study shows the explainable results. It can assist traffic law enforcement agencies in more effectively monitoring and managing the operation of long range coaches, thereby reinforcing codes of practice in the coach industry. 

Future work will focus on further optimizing the algorithm to improve detection accuracy and applicability. Additionally, a deeper analysis of the patterns caused by the different size of road segments will be conducted. Another direction is to incorporate the semi-supervised method to the low rank~\cite{CHAVOSHINEJAD2023-pr}.


 \bibliographystyle{IEEEtran}

\bibliography{refile}

\end{document}